\documentclass{article}
\usepackage[numbers,sort&compress]{natbib}

\usepackage[final]{AdvML_Frontiers_2024}

\usepackage[utf8]{inputenc} 
\usepackage[T1]{fontenc}    
\usepackage{hyperref}       
\usepackage{url}            
\usepackage{booktabs}       
\usepackage{amsfonts}       
\usepackage{nicefrac}       
\usepackage{microtype}      
\usepackage{xcolor}         
\usepackage{graphicx}
\usepackage{amsmath}
\usepackage{algorithm}
\usepackage{algorithmic} 
\usepackage{subcaption}
\usepackage{wrapfig}
\usepackage{multirow}

\title{Adversarial Watermarking for Face Recognition}

\author{
Yuguang Yao \quad\quad Anil Jain\quad\quad Sijia Liu\\
\\
Michigan State University
}

\begin{document}

\maketitle

\begin{abstract}
Watermarking is an essential technique for embedding an identifier (\textit{i.e.}, watermark message) within digital images to assert ownership and monitor unauthorized alterations. In face recognition systems, watermarking plays a pivotal role in ensuring data integrity and security. 
However, an adversary could potentially interfere with the watermarking process, significantly impairing recognition performance.
We explore the interaction between watermarking and adversarial attacks on face recognition models. Our findings reveal that while watermarking or input-level perturbation alone may have a negligible effect on recognition accuracy, the combined effect of watermarking and perturbation can result in an \textit{adversarial watermarking attack}, significantly degrading recognition performance.
Specifically, we introduce a novel threat model, the adversarial watermarking attack, which remains stealthy in the absence of watermarking, allowing images to be correctly recognized initially. However, once watermarking is applied, the attack is activated, causing recognition failures.
Our study reveals a previously unrecognized vulnerability: \textit{adversarial perturbations can exploit the watermark message to evade face recognition systems.} Evaluated on the CASIA-WebFace dataset, our proposed adversarial watermarking attack reduces face matching accuracy by 67.2\% with an $\ell_\infty$ norm-measured perturbation strength
of ${2}/{255}$ and by 95.9\% with a strength of ${4}/{255}$.
\end{abstract}
\vspace{-4mm}
\section{Introduction}
\vspace{-1mm}
Face recognition systems have become increasingly prevalent in various domains, such as access control and surveillance \cite{anwar2020masked,cheng2018surveillance,nasution2020face}. Ensuring the integrity and ownership of facial images used for training and evaluation in such systems is crucial. Image watermarking has offered a viable solution for proprietary face image protection \cite{jain2003hiding,zhu2018hidden,begum2020digital}.
 Watermarking can embed hidden information (also called `watermark message') in digital faces to assert ownership, authenticate content, and verify data integrity \cite{fernandez2023stable, zhang2024double, pal2012biomedical}.

However, as machine learning (ML) models become more sophisticated, they also become susceptible to adversarial attacks. Adversarial perturbations (also known as evasion attacks) are carefully crafted modifications to input data that deceive ML models without noticeable changes in the image to human observers \cite{goodfellow2014explaining,gong2022reverse,zhao2020learning}. In the context of face recognition, such perturbations can cause recognition errors, leading to security breaches; See the literature review in Section\,\ref{sec: related work}. 

Although watermarking aims to protect and authenticate images, the interaction between watermarking processes and adversarial attacks remains underexplored. 
The  presence of watermarking and adversarial attacks, along with their interaction, has added substantial complexity to evaluation of face recognition systems. 
 Inspired by the above, we address the following question:
\begin{center}
    \textit{\textbf{(Q)} How does watermarking affect the adversarial robustness of face recognition systems, and can adversarial attacks exploit watermarking to even degrade face matching performance?
}
\end{center}

To the best of our knowledge, our work unveils the joint effects of watermarking and adversarial attacks on face recognition models for the first time. We summarize our contributions below.

\begin{figure}[t]
    \centering
    \begin{subfigure}[t]{0.7\linewidth}
        \centering
        \includegraphics[width=\linewidth]{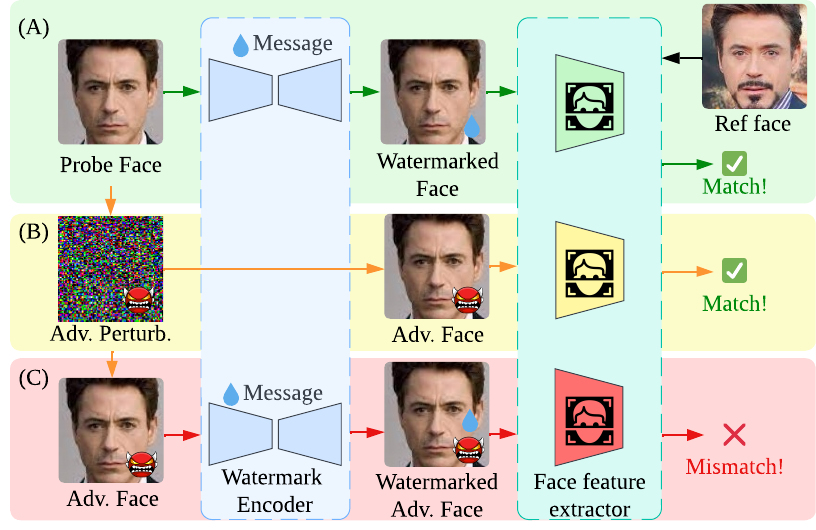}
    \end{subfigure}%

    \caption{\textbf{Overview of the Adversarial Watermarking Attack on Face Recognition.} The green path (A) represents the standard watermarking and face recognition process, where the probe face is watermarked using the watermark encoder and correctly matched with the reference face after feature extraction. The yellow path (B) shows input-level adversarial perturbations applied to evade the face recognition system without watermarking. Subtle adversarial perturbations are added to the probe face, but they do not affect the recognition result without watermarking. The red path (C) demonstrates the adversarial watermarking process, where the adversarially perturbed face image, after being watermarked, fails to match the reference face. 
}
    \label{fig: overview}
    \vspace{-6mm}
\end{figure}

\noindent $\bullet$ We propose a testbed  (Figure\,\ref{fig: overview})  that integrates watermarking techniques into face recognition systems. This framework embeds watermarks into facial images to assert ownership while facilitating the study of adversarial attacks (Figure\,\ref{fig: overview}-(B) and (C)).

\noindent $\bullet$ We introduce a new threat model (Figure\,\ref{fig: overview}-(C)) called the Adversarial Watermarking attack, which differs from conventional evasion attacks against image classifiers \cite{goodfellow2014explaining,szegedy2013intriguing,madry2018towards}. This attack is designed to remain stealthy when watermarking is absent (Figure\,\ref{fig: overview}-(B)), allowing images to be correctly recognized initially. However, once watermarking is applied, the attack is triggered, causing recognition failures and exposing a critical vulnerability in the watermarking process.


\noindent $\bullet$ We validate our proposed attack through extensive experiments on the open-source CASIA-WebFace dataset. Our results demonstrate a significant degradation in face matching performance under small adversarial perturbations (\textit{e.g.}, $\frac{2}{255}$  and  $\frac{4}{255}$) when the watermarking is applied (Figure\,\ref{fig: overview}).

\vspace{-2mm}
\section{Related Work}
\vspace{-1mm}
\label{sec: related work}
\textbf{Watermarking in Face Recognition.} Watermarking techniques have long been used to embed imperceptible information into digital images for purposes such as copyright protection, authentication, and integrity verification \cite{hartung1999multimedia, cayre2005watermarking, begum2020digital}. In the realm of face recognition, watermarking serves as a tool to protect personally identifiable images from unauthorized use and tampering \cite{jain2003hiding, yao2024hide, isa2017watermarking, vatsa2006robust, abdullah2016framework, bhatnagar2013biometrics}. Various methods have been proposed to integrate watermarking into facial images without significantly affecting recognition performance. Traditional watermarking approaches use frequency domain transformations such as Discrete Cosine Transform (DCT) and Discrete Wavelet Transform (DWT) to embed watermarks in images, with the aim of robustness against common image processing attacks \cite{cox2007digital, singh2017dwt}. In contrast, recent methods leverage deep neural networks (DNNs) for watermarking, such as the HiDDeN framework, which employs end-to-end trainable networks to embed and extract watermarks, enhancing resilience against various attacks \cite{zhu2018hidden}. Other recent studies have focused on ensuring that the watermarking process preserves critical facial features essential for accurate recognition \cite{zhang2024double, fernandez2023stable, yao2024hide}. However, these methods mainly focus on robustness against non-adversarial distortions and fail to account for the impact of adversarial perturbations specifically designed to deceive ML models, particularly when watermarking is applied.

\textbf{Adversarial Attacks in Face Recognition.} Adversarial attacks involve introducing subtle, often imperceptible perturbations to input data with the intent of deceiving ML models \cite{szegedy2013intriguing, goodfellow2014explaining,zhang2022robustify}. In face recognition systems, adversarial examples can lead to recognition errors, impersonation, or evasion, posing significant security risks \cite{sharif2016accessorize, dong2019efficient, zhong2020towards}. For example, attackers can manipulate facial images to bypass authentication systems or to impersonate other enrolled individuals in the system. Various attack generation algorithms, such as Fast Gradient Sign Method (FGSM) \cite{goodfellow2014explaining} and Projected Gradient Descent (PGD) \cite{madry2018towards}, have been employed to generate adversarial examples against face recognition models. Meanwhile, defense mechanisms such as adversarial training and input pre-processing have been proposed to mitigate these attacks \cite{madry2018towards,zhang2022robustify,zhang2022revisiting}. The ongoing arms race between attack and defense persists.
However, existing studies have primarily focused on evading or improving the robustness of model performance, without considering the impact of watermarking whose use is growing, \textit{e.g.,} for labeling computer generated images. To the best of our knowledge, the interaction between adversarial perturbations and watermarking in face recognition is largely unexplored, with no prior work investigating how adversarial attacks leverage watermarking to degrade recognition performance.


\vspace{-2mm}
\section{Methods}
\vspace{-1mm}
\label{sec:method}
\textbf{Watermarking System.}
We start by introducing the technique used for generating watermarked face images and its application in the subsequent face recognition task, as shown in Figure\,\ref{fig: overview}-(A). 
To formalize the watermarking problem, let the input image be denoted as $\mathbf{I} \in \mathbb{R}^{H \times W \times C}$, and a binary watermark message as $\mathbf{m} \in \{0, 1\}^L$ (an $L$-bit digital signature) embedded into the facial images \cite{zhu2018hidden,fernandez2023stable,zhao2023recipe}.
Our goal is to produce a watermarked image $\mathbf{I}_{\mathrm{w}}$ that maintains visual similarity to the original image $\mathbf{I}$ containing the watermark message $\mathbf{m}$. Furthermore, the watermarked image should allow extraction of $\mathbf{m}$, allowing provenance of the image.

\begin{wraptable}{r}{0.5\textwidth}
\centering
\vspace{-4mm}
\caption{
{
The robustness of watermarking evaluated using the reconstructed watermark bit accuracy (\%) against various (post-watermarking) data transformations at different scaling strengths. Each value is averaged over 1000 face images, with an image size of 112 $\times$ 112 and a watermark string bit length of 48. See more setup details in Section\,\ref{sec: exp}.
}
}
\resizebox{\linewidth}{!}{%
\begin{tabular}{lcccccc}
\toprule[1pt]
\midrule
\multirow{2}{*}{\textbf{Transformation}} & \multicolumn{6}{c}{\textbf{Scaling ratio}}\\
\cmidrule(lr){2-7}
& 1 & 0.95 & 0.9 & 0.85 & 0.8 & 0.75\\
\midrule
Crop & 98.39 & 97.22 & 93.7 & 95.12 & 94.77 & 94.3\\
Resize & 98.39 & 92.47 & 92.0 & 91.58 & 89.62 & 85.93\\
\midrule
\multirow{2}{*}{\textbf{Transformation}} & \multicolumn{6}{c}{\textbf{Scaling factor}}\\
\cmidrule(lr){2-7}
& 1 & 1.5 & 2 & 2.5 & 3 & 3.5\\
Brightness & 98.39 & 98.48 & 96.65 & 94.21 & 91.6 & 88.87\\
Contrast & 98.39 & 98.81 & 98.15 & 96.82 & 94.92 & 92.62\\
\midrule
\multirow{2}{*}{\textbf{Transformation}} & \multicolumn{6}{c}{\textbf{JPEG quality factor}}\\
\cmidrule(lr){2-7}
& 100 & 95 & 90 & 85 & 80 & 75\\
JPEG compression & 98.39 & 90.36 & 85.0 & 80.8 & 76.65 & 73.06\\
\midrule
\bottomrule[1pt]
\end{tabular}
}
\label{tab: robustness_transformation}
\vspace{-5mm}
\end{wraptable}
We implement the watermarking system using the open source neural network-based HiDDeN framework \cite{zhu2018hidden}. This system consists of an encoder network $f_{\boldsymbol{\theta}}$ and a decoder network $g_{\boldsymbol{\phi}}$. The encoder takes the input image $\mathbf{I}$ and the watermark message $\mathbf{m}$ as inputs and generates the watermarked image $\mathbf{I}_{\mathrm{w}} = f_{\boldsymbol \theta}(\mathbf{I}, \mathbf{m})$. The decoder takes the watermarked image $\mathbf{I}_{\mathrm{w}}$ as input and reconstructs the embedded watermark message $\hat{\mathbf{m}} = g_{\boldsymbol \phi}(\mathbf{I}_{\mathrm{w}})$.
The encoder and decoder networks are jointly trained using a combination of image reconstruction loss and message decoding loss. The loss of image reconstruction $\ell_{\mathrm{recons}}$ (\textit{e.g.}, mean squared error) ensures that the watermarked image is visually similar to the original, while the loss of message decoding $\ell_{\mathrm{decode}}$ (\textit{e.g.}, bitwise binary cross-entropy loss) minimizes the difference between embedded and extracted watermark messages. The overall training objective for watermarking encoder and decoder is:
\begin{align}
\min_{\boldsymbol{\theta}, \boldsymbol{\phi}}  \mathbb{E}_{\mathbf{I}, \mathbf{m}} \left[ \ell_{\mathrm{recons}}(\mathbf{I}_{\mathrm{w}}, \mathbf{I}) + \lambda  \ell_{\mathrm{decode}}(\hat{\mathbf{m}}, \mathbf{m}) \right]
\label{eq: prob_watermark}
\end{align}
where $\lambda$ is a regularization parameter balancing the two losses. 
During training, a random message generator produces random bits for $\mathbf{m}$. This randomness allows the network to generalize to any watermark message, enabling us to embed user-defined messages in face images later on.
Table\,\ref{tab: robustness_transformation} shows that our watermarking system is fairly robust against different data transformations. However, as demonstrated later, this  does not guarantee adversarial robustness for the downstream task when using watermarked data.

\textbf{Face Recognition on Watermarked Images.} With watermarked face images acquired above, we proceed to face recognition to assess the impact of the watermarking. 
In what follows, we provide a brief background on face recognition. 
Given an input face image $\mathbf{I}$, the face recognition model $h_{\boldsymbol{\psi}}$ maps the image to a feature representation $\mathbf{z}$:
$\mathbf{z} = h_{\boldsymbol{\psi}}(\mathbf{I})$,
where $\boldsymbol{\psi}$ represents the learnable parameters of the model. The feature $\mathbf{z}$ is typically extracted from the penultimate layer of a convolutional neural network (CNN), such as ResNet \cite{he2016deep}.
The model is trained to minimize a classification loss, such as the softmax loss \cite{deng2019arcface} or margin-based losses \cite{liu2017sphereface, wang2018cosface, deng2019arcface}, which encourage facial features from the same identity to be close in the embedding space while pushing apart facial features from different identities.
During inference, the model extracts feature representations for a probe face $\mathbf{I}_{\mathrm{p}}$ and a reference face $\mathbf{I}_{\mathrm{r}}$, denoted as $\mathbf{z}_{\mathrm{p}}$ and $\mathbf{z}_{\mathrm{r}}$, respectively. The similarity between the probe and reference faces is computed using the cosine similarity: 
\begin{align}
s(\mathbf{z}_{\mathrm{p}}, \mathbf{z}_{\mathrm{r}}) = \frac{\mathbf{z}_{\mathrm{p}}^\top \mathbf{z}_{\mathrm{r}}}{|\mathbf{z}_{\mathrm{p}}|  |\mathbf{z}_{\mathrm{r}}|}
\label{eq: similarity}
\end{align}
where $|\cdot|$ denotes the Euclidean norm. A match is determined based on whether the similarity score exceeds a predefined threshold $\tau$:
\begin{align}
\text{match}(\mathbf{z}_{\mathrm{p}}, \mathbf{z}_{\mathrm{r}}) =
\begin{cases}
1, & \text{if } s(\mathbf{z}_{\mathrm{p}}, \mathbf{z}_{\mathrm{r}}) \geq \tau, \\
0, & \text{otherwise}.
\end{cases}
\label{eq: match}
\end{align}
Our experiments later verify that the watermarking process does not significantly degrade face recognition performance in the absence of adversarial perturbations.

\textbf{Adversarial Watermarking Attack for Face Recognition.} We introduce an adversarial watermarking attack that exploits the interaction between adversarial perturbations and the watermarking process to degrade face recognition performance.
The adversary aims to craft a minimal perturbation \( \boldsymbol{\delta} \) added to a probe face image \( \mathbf{I}_{\mathrm{p}} \) and find a specific watermark message \( \mathbf{m} \in \{0,1\}^L \) such that:
\vspace{-3mm}
\begin{enumerate}
    \item \textbf{Pre-watermark recognition success:} The perturbed image \( \mathbf{I}_{\mathrm{p}}' = \mathbf{I}_{\mathrm{p}} + \boldsymbol{\delta} \) is correctly matched with the reference image \( \mathbf{I}_{\mathrm{r}} \) by the face recognition model \( h_{\boldsymbol{\psi}} \), \textit{i.e.}, the similarity between their feature representations remains high. Here \( \boldsymbol{\delta} \in \mathbb{R}^{H \times W \times C} \) denotes adversarial perturbations bounded by \( \|\boldsymbol{\delta}\|_\infty \leq \epsilon \), where \( \epsilon \) is  the perturbation strength ensuring imperceptibility. 
    \item \textbf{Post-watermark recognition failure:} After applying the watermarking encoder \( f_{\boldsymbol{\theta}} \) with the adversary-learned  watermark message \( \mathbf{m} \), the  perturbed input image $\mathbf{I}_{\mathrm{p}}'$ and its watermarked counterpart \( \mathbf{I}_{\mathrm{w}}' = f_{\boldsymbol{\theta}}(\mathbf{I}_{\mathrm{p}}', \mathbf{m}) \) lead to  a low similarity with the reference image $\mathbf{I}_{\mathrm{r}}$, causing the face recognition model $h_{\boldsymbol{\psi}}$ to fail.
\end{enumerate}
\vspace{-3mm}
Our rationale has two key aspects. First, satisfying both conditions 1 and 2 ensures that the adversarial attack ($\boldsymbol{\delta}$) stays stealthy when watermarking is absent, but is triggered upon watermark application, leading to recognition failures. Second, this design reveals a unique adversarial challenge in face recognition with watermarking, where the optimization of the watermark message in condition 2 interacts synergistically with the input perturbations $\boldsymbol{\delta}$ to amplify the adversarial effect.

We propose the following  joint optimization problem to find the adversarial perturbation \( \boldsymbol{\delta} \) and the watermark message \( \mathbf{m} \): 
\begin{align}
\min_{\mathbf{m}\in\{0,1\}^L} \,
\min_{\|\boldsymbol{\delta}\|_\infty \leq \epsilon} \,
    - s(\mathbf{z}_{\mathrm{p}}', \mathbf{z}_{\mathrm{r}}) + s(\mathbf{z}_{\mathrm{w}}', \mathbf{z}_{\mathrm{r}})
    \label{eq:adv_attack_with_m}
\end{align}
where the optimization variables are the binary watermark message $\mathbf{m}$ and the input perturbations $\boldsymbol{\delta}$, and
$s(\cdot, \cdot)$ and $\mathbf{z}_{\mathrm{r}}$ are defined in \eqref{eq: similarity}.
Recall that 
\( \mathbf{z}_{\mathrm{p}}' = h_{\boldsymbol{\psi}}(\mathbf{I}_{\mathrm{p}}') \) and \( \mathbf{z}_{\mathrm{w}}' = h_{\boldsymbol{\psi}}( \mathbf{I}_{\mathrm{w}}' ) \) are the feature representations given the probe image $\mathbf{I}_{\mathrm{p}}' = \mathbf{I}_{\mathrm{p}} + \boldsymbol{\delta}$ and $ \mathbf{I}_{\mathrm{w}}' = f_{\boldsymbol{\theta}}(\mathbf{I}_{\mathrm{p}}', \mathbf{m})$, respectively.
In \eqref{eq:adv_attack_with_m}, the original similarity term $s(\mathbf{z}_{\mathrm{p}}', \mathbf{z}_{\mathrm{r}})$ 
ensures that the perturbed face is still recognized as the same identity in the absence of watermarking. And the watermarked similarity term $s(\mathbf{z}_{\mathrm{w}}', \mathbf{z}_{\mathrm{r}})$ minimizes the similarity between the watermarked, perturbed image and the reference image, causing face recognition failure post-watermarking.

To solve the optimization in \eqref{eq:adv_attack_with_m},
we then adopt an alternative optimization procedure    to jointly optimize \( \boldsymbol{\delta} \) and \( \mathbf{m} \).  Specifically, we use the PGD (projected gradient descent) method \cite{madry2018towards} to iteratively minimize one variable while keeping the other fixed.
In the optimization process, we face the challenge of the discrete nature of the watermark message $\mathbf{m}$. 
Direct optimization over binary variables is computationally intractable for large dimensionality $L$. To address this, we relax $\mathbf{m}$ to be continuous in the range $[0,1]^L$ during the optimization. This relaxation allows us to employ PGD in an efficient way.
That is,
after performing gradient descent on the relaxed $\mathbf{m}$, we project back onto the binary set $\{0,1\}^L$ by rounding each element to $0$ or $1$. This ensures the watermark message remains valid for the encoder. By alternately optimizing over $\boldsymbol{\delta}$ and $\mathbf{m}$, we minimize the joint objective. This approach finds a combination of adversarial perturbation and a watermark message that maintains high genuine similarity before watermarking and cause misrecognition afterward.

\vspace{-2mm}
\section{Experiments}
\vspace{-1mm}
\label{sec: exp}
\textbf{Experimental Setup.} We use the \textbf{CASIA-WebFace} dataset \cite{yi2014learning}, containing face images of 10,575 individuals, for evaluating face recognition models. 
We extract 1,000 individuals with two matching face images for each identity ($\mathbf{I}_{\mathrm{p}}$ and $\mathbf{I}_{\mathrm{r}}$), and pre-processed them by aligning and resizing the images to $112 \times 112$ pixels.
We adopt our face recognition model from the    \textbf{AdaFace} framework  \cite{kim2022adaface}. AdaFace is known for its adaptive margin loss that accounts for the quality of the face images, improving recognition performance. The model is trained on MS-Celeb-1M dataset \cite{deng2019arcface} using standard training protocols with a ResNet-50 backbone \cite{he2016deep}. For watermarking, we follow the \textbf{HiDDeN} framework \cite{zhu2018hidden} to solve the problem \eqref{eq: prob_watermark}. The encoder and decoder networks are trained on the MS-COCO dataset \cite{lin2014microsoft} with random 48-bit watermark messages. The trained encoder is then used to embed watermarks in the CASIA-WebFace face images.
%
In generating the adversarial watermarking attack \eqref{eq:adv_attack_with_m},   the step sizes for optimizing $\boldsymbol{\delta}$ and $\mathbf{m}$ are set to $\alpha = \frac{\epsilon}{T}$ and $\beta = \frac{1}{T}$, respectively, where $T=10$ represents the number of iterations for the PGD-10 attack.

\textbf{Evaluation.}
We assess the effectiveness of the adversarial watermarking attack by analyzing face recognition performance under two key conditions. First, in the case of recognition with adversarial perturbations, adversarial perturbations are applied to the probe images \textit{without watermarking}. 
Next, in the case of recognition with the adversarial watermarking attack (\textit{with watermarking}), both adversarial perturbations and an optimized watermark message are applied, following the joint optimization in   \eqref{eq:adv_attack_with_m}.

\begin{figure}[htb]
    \centering
    \vspace{-3mm}
    \includegraphics[width=0.8\linewidth]{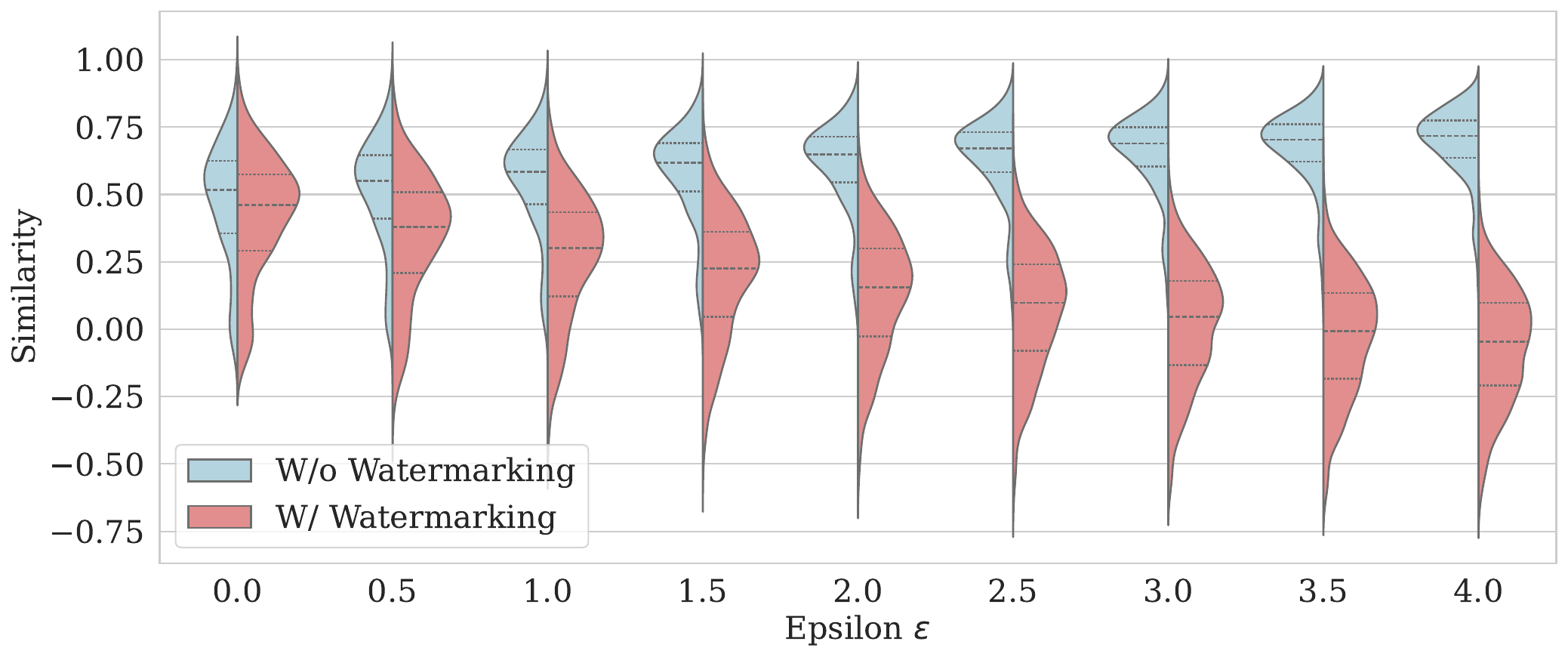}
    \vspace{-3mm}
    \caption{Violin plots of similarity scores in \eqref{eq: similarity} at different $\epsilon$ values (scaled by ${1}/{255}$). For each $\epsilon$, the violin plot shows the distribution of similarity scores between perturbed probe and reference images under two conditions: with watermarking (blue) and without watermarking (red). By $\|\delta_\infty\| \leq \epsilon$, we change $\epsilon$ to control the perturbation strength.
    }
    \label{fig:similarity_violin_plot}
    \vspace{-8mm}
\end{figure}

\begin{table}[htb]
\centering
\caption{Face matching accuracy (\%) with and without watermarking at different perturbation levels ($\epsilon$, scaled by ${1}/{255}$), where the matching threshold is set to $\tau = 0.3$ in \eqref{eq: match}. Performance reduction by watermarking attack is highlighted in blue.}
\label{tab:avg_similarity}
\resizebox{0.8\textwidth}{!}{%
\begin{tabular}{lccccccccc}
\toprule
\textbf{$\epsilon$} & \textbf{0.0} & \textbf{0.5} & \textbf{1.0} & \textbf{1.5} & \textbf{2.0} & \textbf{2.5} & \textbf{3.0} & \textbf{3.5} & \textbf{4.0} \\
\midrule
W/o Watermarking  & 81.8  & 85.4  & 88.5  & 90.9  & 92.2  & 94.1  & 95.7  & 97.5  & 98.3  \\
W/ Watermarking   & 73.9  & 63.5  & 50.0  & 35.7  & 25.0  & 16.5  & 8.4   & 4.5   & 2.4   \\
\textcolor{blue}{Reduction} & \textcolor{blue}{7.9} & \textcolor{blue}{21.9} & \textcolor{blue}{38.5} & \textcolor{blue}{55.2} & \textcolor{blue}{67.2} & \textcolor{blue}{77.6} & \textcolor{blue}{87.3} & \textcolor{blue}{93.0} & \textcolor{blue}{95.9} \\
\bottomrule
\end{tabular}
}
\end{table}

\textbf{Adversarial Watermarking: Joint Effects of Watermarking and Adversarial Perturbations.}
To analyze the effect of the adversarial watermarking attack on face recognition performance, we first examine the similarity scores between probe and reference images across different perturbation strengths $\epsilon$.  \textbf{Figure~\ref{fig:similarity_violin_plot}} 
shows violin plots of the similarity distributions for face recognition, both with and without watermarking, when evaluated using input perturbations $\boldsymbol{\delta}$ from the proposed adversarial watermarking attack. As the perturbation strength $\epsilon$ increases, the similarity between probe and reference images decreases significantly in the presence of watermarking, while it remains largely unaffected without watermarking.
This is because in the absence of watermarking, the first loss term in \eqref{eq:adv_attack_with_m}
 aims to maximize the similarity between the probe image and the reference image for the applied perturbations $\boldsymbol{\delta}$.
With watermarking in the face recognition process, the similarity score quickly drops with increased perturbation strength.  In fact, when $\epsilon = 0.5/255$, the similarity has tended to be smaller than the matching threshold $\tau$ (commonly set at $\tau = 0.3$). This shows that even a small adversarial perturbation can disrupt face recognition after watermarking, although performance remains stable without watermarking.


\textbf{Table~\ref{tab:avg_similarity}} shows that watermarking reduces face matching accuracy at all perturbation levels ($\epsilon$). For example, at $\epsilon = 0.0$, accuracy drops by \textbf{7.9\%} from 81.8\% to 73.9\% after watermarking. This indicates that the adversarial watermark message alone, as found by \eqref{eq:adv_attack_with_m}, reduces recognition accuracy.
As the perturbation magnitude $\epsilon$ increases, the accuracy reduction intensifies. At $\epsilon = {2}/{255}$, the accuracy decreases by \textbf{67.2\%}, from 92.2\% to 25.0\%, and at $\epsilon = {4}/{255}$, the reduction reaches \textbf{95.9\%}, with the accuracy dropping from 98.3\% to just 2.4\%. These drastic reductions illustrate the adversarial watermarking attack's effectiveness in significantly degrading face recognition, especially at higher perturbation magnitudes.
The results demonstrate that adversarial watermarking exploits the interaction with perturbations, significantly reducing face matching accuracy.
a

\begin{figure}[ht]
    \centering
    \vspace{-3mm}
\begin{subfigure}[b]{0.119\linewidth}
    \centering
    \small{Ref}
    \includegraphics[width=\linewidth]{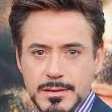}
    \small{Ref}
    \includegraphics[width=\linewidth]{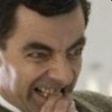}
    \small{Ref}
    \includegraphics[width=\linewidth]{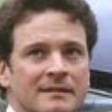}
    \small{Ref}
    \includegraphics[width=\linewidth]{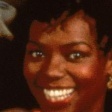}
    \caption{$\mathbf{I}_{\mathrm{r}}$}
\end{subfigure}
\begin{subfigure}[b]{0.119\linewidth}
    \centering
    \small{0.730}
    \includegraphics[width=\linewidth]{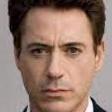}
    \small{0.458}
    \includegraphics[width=\linewidth]{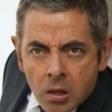}
    \small{0.561}
    \includegraphics[width=\linewidth]{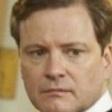}
    \small{0.385}
    \includegraphics[width=\linewidth]{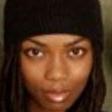}
    \caption{$\mathbf{I}_{\mathrm{p}}$}
\end{subfigure}
\begin{subfigure}[b]{0.119\linewidth}
    \centering
    \small{0.719}
    \includegraphics[width=\linewidth]{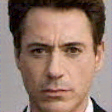}
    \small{0.457}
    \includegraphics[width=\linewidth]{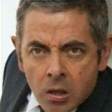}
    \small{0.531}
    \includegraphics[width=\linewidth]{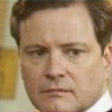}
    \small{0.391}
    \includegraphics[width=\linewidth]{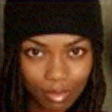}
    \caption{$\mathbf{I}_{\mathrm{w}}$}
\end{subfigure}
\begin{subfigure}[b]{0.119\linewidth}
    \centering
    \small{Diff}
    \includegraphics[width=\linewidth]{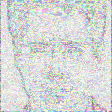}
    \small{Diff}
    \includegraphics[width=\linewidth]{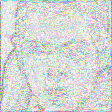}
    \small{Diff}
    \includegraphics[width=\linewidth]{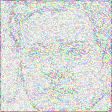}
    \small{Diff}
    \includegraphics[width=\linewidth]{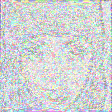}
    \caption{$|\mathbf{I}_{\mathrm{w}} - \mathbf{I}_{\mathrm{p}}|$}
\end{subfigure}
\begin{subfigure}[b]{0.119\linewidth}
    \centering
    \small{0.788}
    \includegraphics[width=\linewidth]{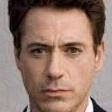}
    \small{0.656}
    \includegraphics[width=\linewidth]{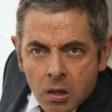}
    \small{0.783}
    \includegraphics[width=\linewidth]{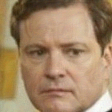}
    \small{0.677}
    \includegraphics[width=\linewidth]{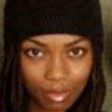}
    \caption{$\mathbf{I}_{\mathrm{p}}'$}
\end{subfigure}
\begin{subfigure}[b]{0.119\linewidth}
    \centering
    \small{Diff}
    \includegraphics[width=\linewidth]{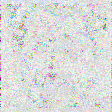}
    \small{Diff}
    \includegraphics[width=\linewidth]{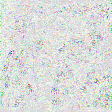}
    \small{Diff}
    \includegraphics[width=\linewidth]{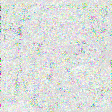}
    \small{Diff}
    \includegraphics[width=\linewidth]{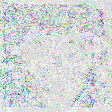}
    \caption{$\boldsymbol{\delta}$}
\end{subfigure}
\begin{subfigure}[b]{0.119\linewidth}
    \centering
    \small{0.169}
    \includegraphics[width=\linewidth]{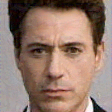}
    \small{0.081}
    \includegraphics[width=\linewidth]{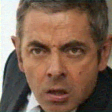}
    \small{0.045}
    \includegraphics[width=\linewidth]{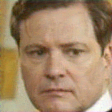}
    \small{-0.057}
    \includegraphics[width=\linewidth]{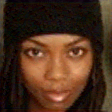}
    \caption{$\mathbf{I}_{\mathrm{w}}'$}
\end{subfigure}
\begin{subfigure}[b]{0.119\linewidth}
    \centering
    \small{Diff}
    \includegraphics[width=\linewidth]{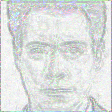}
    \small{Diff}
    \includegraphics[width=\linewidth]{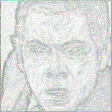}
    \small{Diff}
    \includegraphics[width=\linewidth]{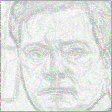}
    \small{Diff}
    \includegraphics[width=\linewidth]{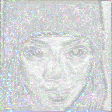}
    \caption{$|\mathbf{I}_{\mathrm{w}}' - \mathbf{I}_{\mathrm{p}}|$}
\end{subfigure}
    \caption{
    Visualization of reference, probe, and perturbed/watermarked face images along with perturbation/watermark for four identities.
    (\textbf{a}) Reference face.
    (\textbf{b}) Probe face.
    (\textbf{c}) Watermarked face.
    (\textbf{d}) Difference between (b) and (c).
    (\textbf{e}) Perturbed face.
    (\textbf{f}) Difference between (b) and (e).
    (\textbf{g}) Adversarial watermarked face by watermarking perturbed face.
    (\textbf{h}) Difference between (b) and (g).
    All element-wise absolute differences are scaled by $\times 10$ and color reverted. All probe faces are marked with their similarity score compared with reference faces at the top of images.
    }
    \label{fig: vis_faces}
    \vspace{-5mm}
\end{figure}

\textbf{Visualizations of Face Images vs. Watermarking and Perturbations.} 
\textbf{Figure\,\ref{fig: vis_faces}} examines the combination of watermarking and perturbations (with strength $\epsilon$ at ${4}/{255}$) on face images.
To compare with reference faces (a), original faces (b) are visualized along with similarity scores by \eqref{eq: similarity}. Watermarked faces (c) are added with message $\mathrm{m}$ by \eqref{eq:adv_attack_with_m}, along with similar scores to (b), exhibiting minor effects by watermarking. Perturbed faces (d) are added with perturbation $\boldsymbol\delta$ by \eqref{eq:adv_attack_with_m}, along with larger scores than (b), maintaining the face matching performance. Adversarial watermarked faces (g) have extremely low similarity scores, exhibiting the joint adversarial effect of watermarking and perturbation. Element-wise absolute differences are visualized in (d), (f), and (h) respectively for (c), (e), (g) to show the imperceptibility of watermark/perturbation. It should be noted that adversarial watermarking difference (h) shows more focus on the edges and corners, \textit{i.e.}, the high frequency area than (d) and (f), illustrating why the attack works while watermakring or perturbation alone does not.



\section{Conclusion}
\vspace{-1mm}

Our study investigated the vulnerabilities of face recognition systems when adversarial perturbations are combined with watermarking. While watermarking alone had a minimal effect on recognition accuracy, the introduction of adversarial perturbations before watermarking caused significant performance degradation. Our findings show that adversarial watermarking attacks could severely undermine recognition systems even if they remain stealthy when watermarking is absent, highlighting the need for improved defenses in both watermarking and face recognition models. 

%

{
\small
  \bibliographystyle{unsrt}
  \bibliography{ref/ref}

\begin{thebibliography}{10}

\bibitem{anwar2020masked}
Aqeel Anwar and Arijit Raychowdhury.
\newblock Masked face recognition for secure authentication.
\newblock {\em arXiv preprint arXiv:2008.11104}, 2020.

\bibitem{cheng2018surveillance}
Zhiyi Cheng, Xiatian Zhu, and Shaogang Gong.
\newblock Surveillance face recognition challenge.
\newblock {\em arXiv preprint arXiv:1804.09691}, 2018.

\bibitem{nasution2020face}
Muhammad Irwan~Padli Nasution, Nurbaiti Nurbaiti, Nurlaila Nurlaila, Tri Inda~Fadhila Rahma, and Kamilah Kamilah.
\newblock Face recognition login authentication for digital payment solution at covid-19 pandemic.
\newblock In {\em 2020 3rd International Conference on Computer and Informatics Engineering (IC2IE)}, pages 48--51. IEEE, 2020.

\bibitem{jain2003hiding}
Anil~K Jain and Umut Uludag.
\newblock Hiding biometric data.
\newblock {\em IEEE transactions on pattern analysis and machine intelligence}, 25(11):1494--1498, 2003.

\bibitem{zhu2018hidden}
Jiren Zhu, Russell Kaplan, Justin Johnson, and Li~Fei-Fei.
\newblock Hidden: Hiding data with deep networks.
\newblock In {\em Proceedings of the European conference on computer vision (ECCV)}, pages 657--672, 2018.

\bibitem{begum2020digital}
Mahbuba Begum and Mohammad~Shorif Uddin.
\newblock Digital image watermarking techniques: a review.
\newblock {\em Information}, 11(2):110, 2020.

\bibitem{fernandez2023stable}
Pierre Fernandez, Guillaume Couairon, Herv{\'e} J{\'e}gou, Matthijs Douze, and Teddy Furon.
\newblock The stable signature: Rooting watermarks in latent diffusion models.
\newblock In {\em Proceedings of the IEEE/CVF International Conference on Computer Vision}, pages 22466--22477, 2023.

\bibitem{zhang2024double}
Yunming Zhang, Dengpan Ye, Sipeng Shen, Caiyun Xie, Ziyi Liu, Jiacheng Deng, and Long Tang.
\newblock Double privacy guard: Robust traceable adversarial watermarking against face recognition.
\newblock {\em arXiv preprint arXiv:2404.14693}, 2024.

\bibitem{pal2012biomedical}
Koushik Pal, G~Ghosh, and M~Bhattacharya.
\newblock Biomedical image watermarking in wavelet domain for data integrity using bit majority algorithm and multiple copies of hidden information.
\newblock {\em American Journal of Biomedical Engineering}, 2(2):29--37, 2012.

\bibitem{goodfellow2014explaining}
Ian~J Goodfellow, Jonathon Shlens, and Christian Szegedy.
\newblock Explaining and harnessing adversarial examples.
\newblock {\em arXiv preprint arXiv:1412.6572}, 2014.

\bibitem{gong2022reverse}
Yifan Gong, Yuguang Yao, Yize Li, Yimeng Zhang, Xiaoming Liu, Xue Lin, and Sijia Liu.
\newblock Reverse engineering of imperceptible adversarial image perturbations.
\newblock {\em arXiv preprint arXiv:2203.14145}, 2022.

\bibitem{zhao2020learning}
Pu~Zhao, Parikshit Ram, Songtao Lu, Yuguang Yao, Djallel Bouneffouf, Xue Lin, and Sijia Liu.
\newblock Learning to generate image source-agnostic universal adversarial perturbations.
\newblock {\em arXiv preprint arXiv:2009.13714}, 2020.

\bibitem{szegedy2013intriguing}
Christian Szegedy, Wojciech Zaremba, Ilya Sutskever, Joan Bruna, Dumitru Erhan, Ian Goodfellow, and Rob Fergus.
\newblock Intriguing properties of neural networks.
\newblock {\em arXiv preprint arXiv:1312.6199}, 2013.

\bibitem{madry2018towards}
Aleksander Madry, Aleksandar Makelov, Ludwig Schmidt, Dimitris Tsipras, and Adrian Vladu.
\newblock Towards deep learning models resistant to adversarial attacks.
\newblock In {\em International Conference on Learning Representations (ICLR)}, 2018.

\bibitem{hartung1999multimedia}
Frank Hartung and Martin Kutter.
\newblock Multimedia watermarking techniques.
\newblock {\em Proceedings of the IEEE}, 87(7):1079--1107, 1999.

\bibitem{cayre2005watermarking}
Fran{\c{c}}ois Cayre, Caroline Fontaine, and Teddy Furon.
\newblock Watermarking security: theory and practice.
\newblock {\em IEEE Transactions on signal processing}, 53(10):3976--3987, 2005.

\bibitem{yao2024hide}
Yuguang Yao, Steven Grosz, Sijia Liu, and Anil Jain.
\newblock Hide and seek: How does watermarking impact face recognition?
\newblock {\em arXiv preprint arXiv:2404.18890}, 2024.

\bibitem{isa2017watermarking}
Mohd Rizal~Mohd Isa, Salem Aljareh, and Zaharin Yusoff.
\newblock A watermarking technique to improve the security level in face recognition systems.
\newblock {\em Multimedia Tools and Applications}, 76:23805--23833, 2017.

\bibitem{vatsa2006robust}
Mayank Vatsa, Richa Singh, Afzel Noore, Max~M Houck, and Keith Morris.
\newblock Robust biometric image watermarking for fingerprint and face template protection.
\newblock {\em IEICE Electronics Express}, 3(2):23--28, 2006.

\bibitem{abdullah2016framework}
Mohammed~AM Abdullah, Satnam~S Dlay, Wai~L Woo, and Jonathon~A Chambers.
\newblock A framework for iris biometrics protection: a marriage between watermarking and visual cryptography.
\newblock {\em IEEE Access}, 4:10180--10193, 2016.

\bibitem{bhatnagar2013biometrics}
Gaurav Bhatnagar and QM~Jonathan Wu.
\newblock Biometrics inspired watermarking based on a fractional dual tree complex wavelet transform.
\newblock {\em Future Generation Computer Systems}, 29(1):182--195, 2013.

\bibitem{cox2007digital}
Ingemar Cox, Matthew Miller, Jeffrey Bloom, Jessica Fridrich, and Ton Kalker.
\newblock {\em Digital watermarking and steganography}.
\newblock Morgan kaufmann, 2007.

\bibitem{singh2017dwt}
Deepika Singh and Satnam Singh.
\newblock Dwt-dct and svd based robust and blind watermarking scheme for copyright protection.
\newblock {\em Multimedia Tools and Applications}, 76(11):13001--13024, 2017.

\bibitem{zhang2022robustify}
Yimeng Zhang, Yuguang Yao, Jinghan Jia, Jinfeng Yi, Mingyi Hong, Shiyu Chang, and Sijia Liu.
\newblock How to robustify black-box ml models? a zeroth-order optimization perspective.
\newblock {\em arXiv preprint arXiv:2203.14195}, 2022.

\bibitem{sharif2016accessorize}
Mahmood Sharif, Sruti Bhagavatula, Lujo Bauer, and Michael~K Reiter.
\newblock Accessorize to a crime: Real and stealthy attacks on state-of-the-art face recognition.
\newblock In {\em Proceedings of the 2016 ACM SIGSAC Conference on Computer and Communications Security (CCS)}, pages 1528--1540. ACM, 2016.

\bibitem{dong2019efficient}
Yinpeng Dong, Hang Su, Baoyuan Wu, Zhifeng Li, Wei Liu, Tong Zhang, and Jun Zhu.
\newblock Efficient decision-based black-box adversarial attacks on face recognition.
\newblock In {\em proceedings of the IEEE/CVF conference on computer vision and pattern recognition}, pages 7714--7722, 2019.

\bibitem{zhong2020towards}
Yaoyao Zhong and Weihong Deng.
\newblock Towards transferable adversarial attack against deep face recognition.
\newblock {\em IEEE Transactions on Information Forensics and Security}, 16:1452--1466, 2020.

\bibitem{zhang2022revisiting}
Yihua Zhang, Guanhua Zhang, Prashant Khanduri, Mingyi Hong, Shiyu Chang, and Sijia Liu.
\newblock Revisiting and advancing fast adversarial training through the lens of bi-level optimization.
\newblock In {\em International Conference on Machine Learning}, pages 26693--26712. PMLR, 2022.

\bibitem{zhao2023recipe}
Yunqing Zhao, Tianyu Pang, Chao Du, Xiao Yang, Ngai-Man Cheung, and Min Lin.
\newblock A recipe for watermarking diffusion models.
\newblock {\em arXiv preprint arXiv:2303.10137}, 2023.

\bibitem{he2016deep}
Kaiming He, Xiangyu Zhang, Shaoqing Ren, and Jian Sun.
\newblock Deep residual learning for image recognition.
\newblock In {\em Proceedings of the IEEE conference on computer vision and pattern recognition}, pages 770--778, 2016.

\bibitem{deng2019arcface}
Jiankang Deng, Jia Guo, Niannan Xue, and Stefanos Zafeiriou.
\newblock Arcface: Additive angular margin loss for deep face recognition.
\newblock In {\em Proceedings of the IEEE/CVF conference on computer vision and pattern recognition}, pages 4690--4699, 2019.

\bibitem{liu2017sphereface}
Weiyang Liu, Yandong Wen, Zhiding Yu, Ming Li, Bhiksha Raj, and Le~Song.
\newblock Sphereface: Deep hypersphere embedding for face recognition.
\newblock In {\em Proceedings of the IEEE conference on computer vision and pattern recognition}, pages 212--220, 2017.

\bibitem{wang2018cosface}
Hao Wang, Yitong Wang, Zheng Zhou, Xing Ji, Dihong Gong, Jingchao Zhou, Zhifeng Li, and Wei Liu.
\newblock Cosface: Large margin cosine loss for deep face recognition.
\newblock In {\em Proceedings of the IEEE conference on computer vision and pattern recognition}, pages 5265--5274, 2018.

\bibitem{yi2014learning}
Dong Yi, Zhen Lei, Shengcai Liao, and Stan~Z Li.
\newblock Learning face representation from scratch.
\newblock {\em arXiv preprint arXiv:1411.7923}, 2014.

\bibitem{kim2022adaface}
Minchul Kim, Anil~K Jain, and Xiaoming Liu.
\newblock Adaface: Quality adaptive margin for face recognition.
\newblock In {\em Proceedings of the IEEE/CVF conference on computer vision and pattern recognition}, pages 18750--18759, 2022.

\bibitem{lin2014microsoft}
Tsung-Yi Lin, Michael Maire, Serge Belongie, James Hays, Pietro Perona, Deva Ramanan, Piotr Doll{\'a}r, and C~Lawrence Zitnick.
\newblock Microsoft coco: Common objects in context.
\newblock In {\em Computer Vision--ECCV 2014: 13th European Conference, Zurich, Switzerland, September 6-12, 2014, Proceedings, Part V 13}, pages 740--755. Springer, 2014.

\end{thebibliography}
}

\end{document}